\documentclass[runningheads]{llncs}
\usepackage{graphicx}
\usepackage{amsmath,amssymb} 
\usepackage{color}
\usepackage{overpic}
\usepackage{multirow}
\usepackage[width=122mm,left=12mm,paperwidth=146mm,height=193mm,top=12mm,paperheight=217mm]{geometry}

\begin{document}
\title{Disassembling Object Representations \\without Labels}

\titlerunning{Disassembling Object Representations without Labels}
%
\author{Zunlei Feng\inst{1} \and
Xinchao Wang\inst{2}\and
Yongming He\inst{1}\and
Yike Yuan\inst{1}\and
Xin Gao\inst{3}\and
Mingli Song\inst{1}
}
%
\authorrunning{Zunlei Feng, Xinchao Wang, et al.}
%

\institute{Zhejiang University
\email{\{zunleifeng,yongminghe,yuanyike,brooksong\}@zju.edu.cn} \and
Stevens Institute of Technology
\email{xinchao.wang@stevens.edu} \and
Alibaba Group
\email{zimu.gx@alibaba-inc.com}}
\maketitle              

\begin{abstract}
In this paper, we study a new representation-learning task, which we termed as disassembling object representations. Given an image featuring multiple objects, the goal of disassembling is to acquire a latent representation, of which each part corresponds to one category of objects. Disassembling thus finds its application in a wide domain such as image editing and few- or zero-shot learning, as it enables category-specific modularity in the learned representations.
To this end, we propose an unsupervised approach to achieving disassembling, named Unsupervised Disassembling Object Representation (UDOR). UDOR follows a double auto-encoder architecture, in which a fuzzy classification and an object-removing operation are imposed. The fuzzy classification constrains each part of the latent representation to encode features of up to one object category, while the object-removing, combined with a generative adversarial network, enforces the modularity of the representations and integrity of the reconstructed image. Furthermore, we devise two metrics to respectively measure the modularity of disassembled representations and the visual integrity of reconstructed images.
Experimental results demonstrate that the proposed UDOR, despited unsupervised, achieves truly encouraging results on par with those of supervised methods.
\end{abstract}

\section{Introduction}
\begin{figure}[!t]
\centering
\includegraphics[scale =0.32]{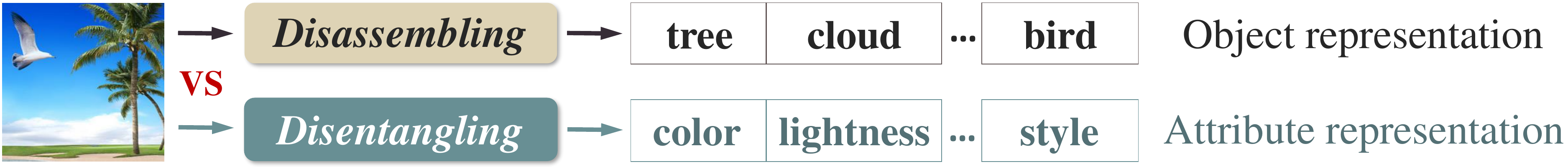}
\caption{Comparing disentangling and the proposed disassembling.
Given an input image, both tasks aim to learn an interpretable representation, of which each part is semantically explainable.
Disentangling focuses on extracting attribute-specific
representation parts,
while the disassembling targets at
deriving object-specific parts.}
\label{disassemblingVSdisentangling}
\end{figure}

Deep learning has led to unprecedented performances in many computer vision and machine learning tasks.
The success of deep networks is largely attributed to their capability to learn representations automatically from the sheer amount of data.
Despite the pleasing results,
the learned representations, especially at a more global level, are in many cases not explainable.
Given a landscape image like the one shown in Fig.~\ref{disassemblingVSdisentangling}, existing feature-learning approaches focus on producing a global representation for the whole image, in which the features of the scene objects, like the tree and bird, are intertwined.
Such object-tangled representations, in many cases, make a computer vision task like image editing cumbersome to be deployed.

We study in this paper a novel task,
termed as \emph{disassembling object representations},
towards  interpretable representation learning.
Unlike prior  disentangling task that
isolates attributes of different natures like color and lightness,
disassembling aims at learning object-specific features.
More specifically, given an image depicting  multiple objects,
disassembling attempts
to separate features of objects from different categories into distinct parts of a latent representation.
In this way, each such part encodes
only the features of objects in one specific category.
In Fig.~\ref{disassemblingVSdisentangling},
for example, one part of the disassembled representation corresponds to the tree, and another corresponds to the bird.




Disassembling  may therefore potentially
serve as an instrumental step for
machine learning tasks, including but not limited to
image editing, few- or zero-shot learning~\cite{akata2015evaluation,song2018transductive,xian2017zero-shot}, and image classification~\cite{nam2014large,read2010scalable}.
In the case of image editing, disassembling makes it possible
to alter the appearance of designated objects via
directly working on their isolated
and thus filtered latent representation.
In the case of few- or zero-shot learning,
disassembling  allows us to potentially extract pure and intact features from foreground objects and meanwhile suppresses the irrelevant ones  from the background.


Towards solving the proposed disassembling task, we propose
an unsupervised approach, which name as
Unsupervised Disassembling Object Representation~(UDOR).
UDOR is motivated by \emph{visual integrity},
referring to the fact that the scene
should remain visually plausible
if one or multiple scene objects are removed.
Given a collection of images,
the UDOR will extract disassembled object representations, of which each part corresponds to one category of objects.

The proposed UDOR comprises three major
components: double AutoEncoder (AE),
Fuzzy Classification, and Object-removing Operation,
as shown in Fig.~\ref{framework}.
UDOR follows a double AE architecture,
which consists of an Image Reconstruction AE~(IR-AE) and
a Representation Reconstruction AE~(RR-AE).
The IR-AE and RR-AE are used to reconstruct the input image and the latent representation, respectively.
The Fuzzy Classification component, on the other hand,
is devised to constrain each part of the latent representation to encode features of up to one object category.
As will be discussed later,
it explicitly accounts for the fact that
features of objects from the same
category should be similar, while
those from different categories should be distinct.
The Object-removing Operation  enforces the modularity of the derived representation.
Specifically, we randomly reset parts of representation to empty vectors, meaning that we remove the corresponding objects from the scene,
and then use a WGAN-GP~\cite{Gulrajani2017Improved} to produce a visually plausible reconstructed image so as to preserve the visual integrity.



To evaluate the disassembling performance, we also propose
two  metrics, one on modularity and the other on integrity.
The former one measures the modularity and portability
of the latent representation, while the latter evaluates the visual quality
of the reconstructed image.
As will be demonstrated in our experiments,
the proposed UDOR, in spite of its unsupervised nature,
achieves truly promising results that closely approach those of the supervised methods.

Our main contributions therefore include introducing
the disassembling task and an unsupervised approach,
termed as UDOR, towards solving it.
UDOR follows a double AE architecture, with
a dedicated Fuzzy Classification component and
an Object-removing Operation combined with a WGAN-GP,
to comply with the modularity of the learned latent representation.
We also introduce two disassembling metrics,
upon which the proposed UDOR achieves truly encouraging results
almost on par with those of the supervised methods.


\section{Related Work}
There are, to our best knowledge, few methods tailored for learning disassembled object representations. The most related works are disentangled representation learning methods, which aim at learning dimension-wise interpretable attribute representation from image data.

Existing disentangling methods can be broadly classified into two categories: unsupervised and supervised approaches. Most of the existing unsupervised methods~\cite{Burgess2017Understanding,Chen2018Isolating,Dupont2018Joint,Gao2018Auto,Kim2018Disentangling} are based on the two most prominent methods \(\beta\)-VAE~\cite{Higgins2016beta} and InfoGAN~\cite{Chen2016InfoGAN}. They impose the independent assumption of different dimensions of the latent representation to achieve disentangling. However, those methods only can disentangle the image's attribute features, such as color, lightness, style, and so on. On the other hand, supervised methods~\cite{Banijamali2017JADE,Feng2018Dual,Kingma2014Semi,Perarnau2016Invertible,Wang2017Tag} focus on utilizing annotated data to supervise the input-to-attribute mapping explicitly. The original aim of the supervised method is to learn disentangled attribute representation. Through annotating the object information as the labels, a few supervised representation disentangling method can also be transferred to learn disassembled object representation. However, it requires a large amount of annotated samples.  There are still some scene decomposition methods~\cite{burgess2019monet:,eslami2016attend,greff2019multi-object,van2018relational}, which also can learn object-related features. However, those methods only can handle toy datasets.

We also give a brief review here about \emph{double AE}, \emph{Fuzzy Classification}, and \emph{Object-removing Operation}, which relate to our UDOR. For the \emph{double AE}, Feng \emph{et al.}~\cite{Feng2018Dual} and Gonzalezgarcia \emph{et al.}~\cite{gonzalezgarcia2018image-to-image} propose the Dual Swapping Disentangling (DSD) and  cross-domain autoencoders to disentangle attribute representation with multiple autoencoders, respectively. The difference between our UDOR and them is that DSD needs two images as input simultaneously, which leads to the fixed framework with four autoencoders.

For the \emph{Fuzzy Classification}, there is no directly related work so far to our knowledge. The most similar methods are some multi-label classification works \cite{Goutsu2018Classification,Rothe2016Deep,wang2016cnn,wu2015weakly}, which transform the multi-label classification into multiple single label classification tasks. However, those methods are all supervised by annotated labels, where the unsupervised fuzzy classification problem does not exist.

For the \emph{Object-removing Operation},
Arandjelovic \emph{et al.}~\cite{arandjelovic2019object} propose the copy-pasting GAN, which copies and pasts object-related parts of an image into a new image. The copy-pasting GAN is devised to learn object mask. Different from the above method, the Object-removing Operation replaces part of the latent representation with an empty vector.
Besides, GAN~\cite{Goodfellow2014Generative} has been applied extensively. In our method, the WGAN-GP~\cite{Gulrajani2017Improved} is adopted to improve the quality of the generated image reconstructed with the object-removed representation.

\section{The Disassembling  Task}
The definition of the disassembling object representation task is given as follows. It is assumed that we are given a dataset that contains $n$ categories of objects. Each sample, in our case taking the form of an image,
is composed of $m$ ($ 0 \leq m \leq n$) categories of objects.
The object of the same category may appear multiple times in one sample. The granularity of the categories is determined by the original dataset and the requirements of the application.

For each sample in the dataset, the disassembled object representation is expected to meet two criteria:
it should contain features of all objects in the sample,
and each part of the disassembled object representation should \emph{only} contain the entire features of the same category of objects in the sample.

\section{The Proposed Method}

In this section, we give more details of our proposed UDOR (Fig.~\ref{framework}). We start by introducing the basic architecture {double AE}, then describe the  {Fuzzy Classification} component, and finally expound the {Object-removing Operation}.

\subsection{Double AE}
\label{DAE}
\begin{figure}[!t]
\centering
\begin{overpic}[scale =0.94]{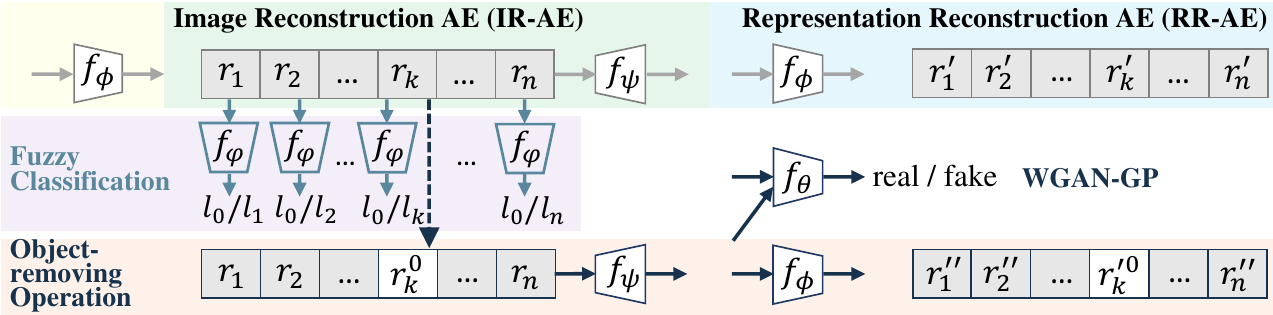}
\put(0.5,18.0){$\mathcal{I}$}
\put(55.4,18.1){$\mathcal{I'}$}
\put(54.4,10.){$\mathbb{P}_{d}$}
\put(54.4,2.4){$\mathcal{I'}^{0}$}

\put(13.0,18.1){$\mathcal{R}$}
\put(12.2,2.1){$\mathcal{R}^0$}
\put(67.8,18.1){$\mathcal{R}'$}
\put(67.8,2.4){$\mathcal{R}''$}

\end{overpic}
\caption{The framework of UDOR. It comprises three components: double AutoEncoder (AE), Fuzzy Classification, and Object-removing Operation. $f_{\phi}$, $f_{\psi}$, $f_{\theta}$, $f_{\varphi}$ and $\mathbb{P}_{d}$ denote the encoder, decoder, discriminator, classifier, and original data distribution, respectively.
The double AE includes an Image Reconstruction AE (IR-AE) and a Representation Reconstruction AE (RR-AE),
which are used to reconstruct the input image $\mathcal{I}$ and
the latent representation $\mathcal{R}$ ($[r_1,r_2,...,r_k,...,r_n]$),
respectively.
The Fuzzy Classification  constrains
that the $k$-th part of the representation,
denoted by $r_{k}$,
only contain features of $k$-th object category~(label $l_{k}$) or the empty features~(label $l_{0}$).
Here, $n$ denotes the number of object categories in the dataset. The Object-removing Operation
replaces a randomly-selected part $r_{k}$ to
an empty vector $r^0_{k}$.
The same RR-AE is adopted to reconstruct the object-removed representation $\mathcal{R}^0$
($[r_1,r_2,...,r^0_k,...,r_n]$), and the WGAN-GP~\cite{Gulrajani2017Improved} is adopted to enhance the quality of the generated image $\mathcal{I'}^{0}$. }
\label{framework}
\end{figure}
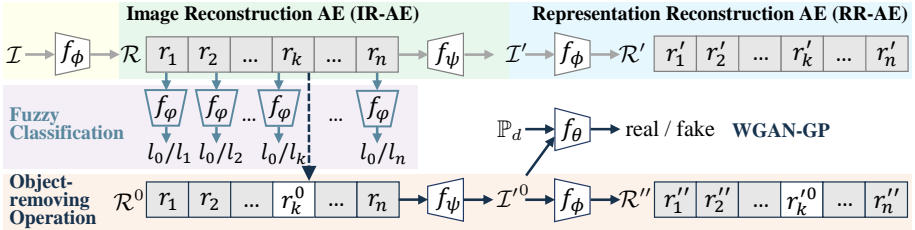

The goal of our proposed UDOR is to train an autoencoder that can disassemble features of different objects in an image into different parts of a latent representation. To get proper initial representation, we adopt a double AE architecture, including an Image Reconstruction AE (IR-AE) and a  Representation Reconstruction AE (RR-AE), shown in Fig.~\ref{framework}. The IR-AE is composed of an encoder $f_{\phi}$ and a decoder $f_{\psi}$. Through reconstructing the input image $\mathcal{I}$, the IR-AE  ensures that the latent representation $\mathcal{R}$ contains all the features of the input image.
With  $\mathcal{R}$ as input, the RR-AE reconstructs the representation $\mathcal{R'}$, which  enhances the  consistency between the input image $\mathcal{I}$ and the representation $\mathcal{R}$. Therefore, the basic reconstruction loss $\mathbf{\mathcal{L}_{rec}}$ is defined as:
\begin{equation}\label{eq1}
\mathbf{\mathcal{L}_{rec}}=||\mathcal{I}-\mathcal{I'}||^2_2+\rho||\mathcal{R}-\mathcal{R'}||^2_2,
\end{equation}
where $\mathcal{R}=f_{\phi}(\mathcal{I})$, $\mathcal{I'}=f_{\psi}(\mathcal{R})$, $\mathcal{R'}=f_{\phi}(\mathcal{I'})$ and $\rho$ is the balance parameter. It is noticeable that all the encoders and decoders share the  same parameters, respectively.

The latent representation $\mathcal{R}$ is devised to be composed of $n$ parts $[r_1,r_2,...,r_k,$ $...,r_n]$, where each part $r_k,k\in\{1,2,3,...,n\}$ is multi-unit.  The hyper-parameter $n$ is decided by the total number of object categories in the image dataset.  By the same token, $\mathcal{R'}$ is also split into $n$ parts $[r'_1,r'_2,...,r'_k,...,r'_n]$.

\subsection{Fuzzy Classification}
 The features of the objects are still intertwined upon the derivation of the initial representation $\mathcal{R}$. We thus propose the Fuzzy Classification component
 to disassemble each object's features into different parts of the representation.
 For the ideal object representation, each part $r_k,k\in\{1,2,3,...,n\}$ should only contain entire features of the same category of objects or empty features. For those nonempty parts, features of objects from the same category should be similar, and those from different categories should be distinct.
 When one category of object is absent from the scene,
 and the corresponding part of representation should be empty. Therefore, it is a \emph{fuzzy classification problem} how to supervise each part of the representation contain features of objects or empty features.

To solve the above fuzzy classification problem, we propose the fuzzy classification loss. For the dataset with $n$ object categories, there are $n+1$ kinds of features (features of $n$ object categories and the empty features $r^0$).
To classify unlabel objects in samples, we predefine the ground-truth label of the representation's $k$-th part with label $l_k$ and label $l_0$. Meanwhile, the ground-truth label of empty features $r^0$ is defined with label $l_0$, which is used to strength the classifier's ability for identifying empty features $r^0$.
For the ideal object representation, the $k$-th part of representation is expected to contain features of the $k$-th object category or the empty features $r^0$. With the $r_k$ as input, the object classifier $f_{\varphi}$ predicts the class probability $p_{k}$, which should be equal to the $k$-th category's label $l_k$ or the label $l_0$. So, the fuzzy classification loss $\mathbf{\mathcal{L}_{cla}}$ is defined as:
\begin{equation}\label{eq2}
\mathbf{\mathcal{L}_{cla}}=-\sum^n_{k=1}\log\{[1-\sum^{vec}(l_0-l_0 \times p_k)(l_k-l_k\times p_k)]\}-\tau\sum^{vec}[ l_0\times \log(p_0)],
\end{equation}
where $l_k$ is one-hot label vector, $p_{k}=f_{\varphi}(r_k)$, $p_0=f_{\varphi}(r^0)$, $\sum^{vec}$ denotes summation of multi-dimension vector, and $\tau$ is the balance parameter.

\subsection{Object-removing Operation}
\label{Object-removing}
With the above Fuzzy Classification, each part of representation $r_k$ will contain relevant features of the specific object. However, irrelevant features of other objects may remain in $r_k$. To enhance the modularity of latent representation, we propose the Object-removing Operation. As described above, when removing some objects, the image remains reasonable and integrated, which is called as \emph{visual integrity}. Base on the visual integrity, we randomly reset part of representation $r_k, k\in\{1,2,3,...,n\}$ to the empty vector $r^0_{k}$, which generates the object-removed representation $\mathcal{R}^0$ ($[r_1,r_2,...,r^0_k,...,r_n]$). The empty vector $r^0_{k}$ is a part of representation extracted from an empty image $\mathcal{I}_{e}$, such as a full-white image, a full-black image, or other kinds of images, which is decided by each training dataset.

If the reset part $r_k$  contains independent and complete features of one object category, the RR-AE should reconstruct the object-removed representation $\mathcal{R}^0$ perfectly. The object-removing loss $\mathbf{\mathcal{L}_{rem}}$ is thus devised to reconstruct the object-removed representation $\mathcal{R}^0$  and the empty image $\mathcal{I}_{e}$ :
\begin{equation}\label{eq4}
\mathbf{\mathcal{L}_{rem}}=||\mathcal{R}^0-\mathcal{R''}||^2_2+\omega||\mathcal{I}_{e}-\mathcal{I}'_{e}||^2_2,
\end{equation}
where $\mathcal{I'}^{0}=f_{\psi}(\mathcal{R}^0)$ is the generated image with object-removed representation $\mathcal{R}^0$, $\mathcal{R}''=f_{\phi}(\mathcal{I'}^{0})$, $\mathcal{I}'_{e}=f_{\psi}(f_{\phi}(\mathcal{I}_{e}))$, and $\omega$ is the balance parameter.



To keep the \emph{visual integrity} of the object-removed image $\mathcal{I'}^{0}$, the WGAN-GP~\cite{Gulrajani2017Improved} is adopted to discriminate that the corresponding objects are removed from $\mathcal{I'}^{0}$. Similar to~\cite{Gulrajani2017Improved}, the generative adversarial loss $\mathbf{\mathcal{L}_{adv}}$ is given as:
\begin{equation}\label{eq3}
\mathbf{\mathcal{L}_{adv}}=\underbrace{\mathop{\mathbb{E}}\limits_{\mathcal{I'}^0 \sim \mathbb{P}_{g}} [f_{\theta}(\mathcal{I'}^0)] - \mathop{\mathbb{E}}\limits_{\mathcal{I} \sim \mathbb{P}_{d}} [f_{\theta}(\mathcal{I})]}_{\text{Original critic loss}} +\underbrace{\lambda \mathop{\mathbb{E}}\limits_{\mathcal{I'}^0 \sim \mathbb{P}_{t}} [(\|\nabla_{\mathcal{I'}^0} f_{\theta}(\mathcal{I'}^0)\|_{2}-1)^2]}_{\text{Gradient penalty}},
\end{equation}
where $\mathbb{P}_{g}$ is the generator distribution, $\mathbb{P}_{d}$ is the original data distribution, $\mathbb{P}_{t}$ is the distribution of generated image $\mathcal{I'}^0$ in training, and $\lambda$ is the balance parameter. The $\mathbf{\mathcal{L}_{adv}}$ will constrain the $k$-th part $r_k$ only contain entire features of the specific category of objects or empty features.

In summary, the total loss $\mathbf{\mathcal{L}}$ contains four parts: the basic reconstruction loss $\mathbf{\mathcal{L}_{rec}}$, fuzzy classification loss $\mathbf{\mathcal{L}_{cla}}$, object-removing loss $\mathbf{\mathcal{L}_{rem}}$ and generative adversarial loss $\mathbf{\mathcal{L}_{adv}}$. $\mathbf{\mathcal{L}_{rec}}$ ensures the features' consistency between the input image and the latent representation. $\mathbf{\mathcal{L}_{cla}}$ ensures that each part of the representation contains features of the specific object category or the empty features. $\mathbf{\mathcal{L}_{rem}}$ is devised to enhance the modularity of latent representation. $\mathbf{\mathcal{L}_{adv}}$ is adopt to improve the quality of the object-removed image reconstructed with the object-removed representation, which can ensure the object-removed image's visual integrity. The total loss $\mathbf{\mathcal{L}}$ is therefore given as follows:
\begin{equation}\label{eq5}
\mathbf{\mathcal{L}}=\alpha\mathbf{\mathcal{L}_{rec}}+\beta\mathbf{\mathcal{L}_{cla}}
+\gamma\mathbf{\mathcal{L}_{rem}}+\eta\mathbf{\mathcal{L}_{adv}},
\end{equation}
where $\alpha$, $\beta$, $\gamma$ and $\eta$ are the balance parameters.

\section{Disassembling Metric}

It is essential to measure the disassembling performance of different methods. To measure the disassembling performance effectively and fairly, we begin by defining the properties that we expect a disassembled representation to have. Then we describe our devised two metrics for quantitatively comparing the disassembling performance.

As described above, the image is usually comprised of several objects, which are removable from the scene.  We therefore assume that the sample in the dataset is generated by a ground truth simulation process that uses different kinds of objects randomly. In this paper, we generate the Multi-MNIST dataset with the handwritten digits \cite{lecun1998gradient-based} as objects. For the $28\times28$ handwritten digit image, we first resize it to $14\times14$ image, and then  create an empty $32\times32$ black image. For the top-left, top-right and left-bottom of the black image, we insert different digit $0$ / nothing, digit $1$ / nothing and digit $2$ / nothing, respectively. As a result, the generated dataset sample may contain nothing or one/ two/ three digits. For the Multi-MNIST dataset, the ideal disassembled representation is that each part of representation will only contain the entire features of a specific digit or empty features. When removing part of representation that contains digit's features, the corresponding digit should disappear completely.

Therefore, we devise two disassembling metrics to measure the \emph{modularity }on the latent representation and the \emph{integrity} on the reconstructed image, respectively. For the \emph{modularity}, we run inference on images that are generated by fixing one object while randomly sampling all other objects or empty. If the modularity property holds for the inferred representations, there will be less variance in the inferred latent representations that correspond to the fixed object.  In this paper, we randomly choose $T$ fixed digits. For the $t$-th fixed digit, $D$ images are generated through random sampling all other digits or empty. Thus, we will get $T\times D$ test images $\{\mathcal{I}^t_{d},t\in\{1,2,3,...,T\},d\in\{1,2,3,...,D\}\}$. For each group of digit-fixed samples, we can get $D$ disassembled digit representation parts $\{z^t_{d},d\in\{1,2,3,...,D\}\}$ that correspond to the $t$-th fixed digit. Then, the \textbf{Modularity Score} $M(T,D)$, which measures the average difference value of $z^t_{d}$, is calculated as follows:
\begin{equation}\label{eq6}
M(T,D)=\frac{1}{T\times D}\sum^T_{t=1}\sum^D_{d=1}\sum^{vec}|z^t_{d}-\frac{1}{D}\sum^D_{d=1}z^t_{d}|,
\end{equation}
where $\sum^{vec}_{1}$ denotes summation of multi-dimension vector. For the \emph{integrity}, we run reconstruction on the same test images $\{\mathcal{I}^t_{d},t\in\{1,2,3,...,T\},d\in\{1,2,3,...,D\}\}$ through resetting the $z^t_{d}$ as an empty vector, which will reconstruct the fixed number disappeared images $\{\widehat{\mathcal{I}}^t_{d},t\in\{1,2,3,...,T\},d\in\{1,2,3,...,$ $ D\}\}$. Meanwhile, the corresponding ground-truth images $\{\overline{\mathcal{I}}^t_{d},t\in\{1,2,3,...,T\},$ $d\in\{1,2,3,...,D\}\}$ are generated through replacing the fixed digit with a $16\times16$ black patch. Then, the \textbf{Integrity Score} $V(T,D)$, which measures the visual integrity of reconstructed images, is defined as follows:
\begin{equation}\label{eq7}
V(T,D)=\frac{1}{T\times D\times W}\sum^T_{t=1}\sum^D_{d=1}\sum^{W}|\overline{\mathcal{I}}^t_{d}-\widehat{\mathcal{I}}^t_{d}|,
\end{equation}
where $W$ is the pixel number of the image, $\sum^{W}$ denotes summation of image pixel difference value.

\section{Experiments}
In this section, we first introduce implementation details and compare the results of our UDOR and other methods qualitatively. Then, we adopt the \emph{Modularity Score} and the \emph{Integrity Score} to evaluate the performance of different methods quantitatively. Next, we give the experiment how the object position influences the disassembling object representation. Meanwhile, we do the ablation study that verifies the effects of different loss terms. What's more, we demonstrate the application performance on the image editing and image classification task.  Lastly, some failure cases are expounded, and some interesting and challenging directions are discussed. (\emph{More results, experimental details, and source codes are given in the supplementary material}.)

\subsection{Experimental Setting}

\textbf{Dataset}. In the experiment, we compare UDOR with other methods on six datasets.
For \emph{Multi-MNIST}, each sample is generated through randomly fill the downsampled handwritten images of digit zero, one, and two or a $16\times16$ empty black patch into the top-left, top-right, and left-bottom of a $32\times32$ black image, respectively.
For \emph{Pattern-Design}, we firstly collect a basic pattern and cut the basic part, which can recombine the pattern through jointing repetitively. The basic part contains three kinds of objects: the flower and the leaf and the tree. Then, we resize the basic part into the $64\times64$ image. For the resized basic part, we generate $50,000$ basic parts through rendering objects with different colors.
For \emph{Multi-Fashion}, each sample is generated by choosing and combining some t-shirt, pants, bag, and shoes from Fashion-MNIST \cite{xiao2017fashion-mnist:} randomly. The position for the t-shirt, pants, bag, and shoes are top-left, left-bottom, top-right, and right-bottom of the generated $64\times64$ outfit image, respectively. In the outfit, some clothing items are allowed nonexistent, which satisfies the real-life scenes.
For \emph{Mugshot}~\cite{Feng2018Dual},  the dataset contains selfie images of different subjects with different backgrounds. What's more, some selfie images with the white background are added into the training datasets. Finally, there is a total of $30,000$ samples in the dataset.
For \emph{Outfit}~\cite{feng2018interpretable}, each sample is generated with an outfit composition algorithm with real clothing items as input. There are up to five clothing items in each outfit image. The outfit dataset contains $20,000$ samples.
The \emph{HAM}~\cite{tschandl2018the} is a large collection of multi-source dermatoscopic images of common pigmented skin lesions. There are $10,015$ dermatoscopic images in the dataset.

\textbf{Network Architectures}. In the experiment, we adopt two kinds of network architecture for image size $32\times32$ and $64\times64$, respectively. The encoders and decoders of $64\times64$ network architecture have the same architecture as ResNet~\cite{eastwood2018a}. More details are given in the supplementary material B.


\begin{figure}[!t]
\centering
\includegraphics[scale =0.85]{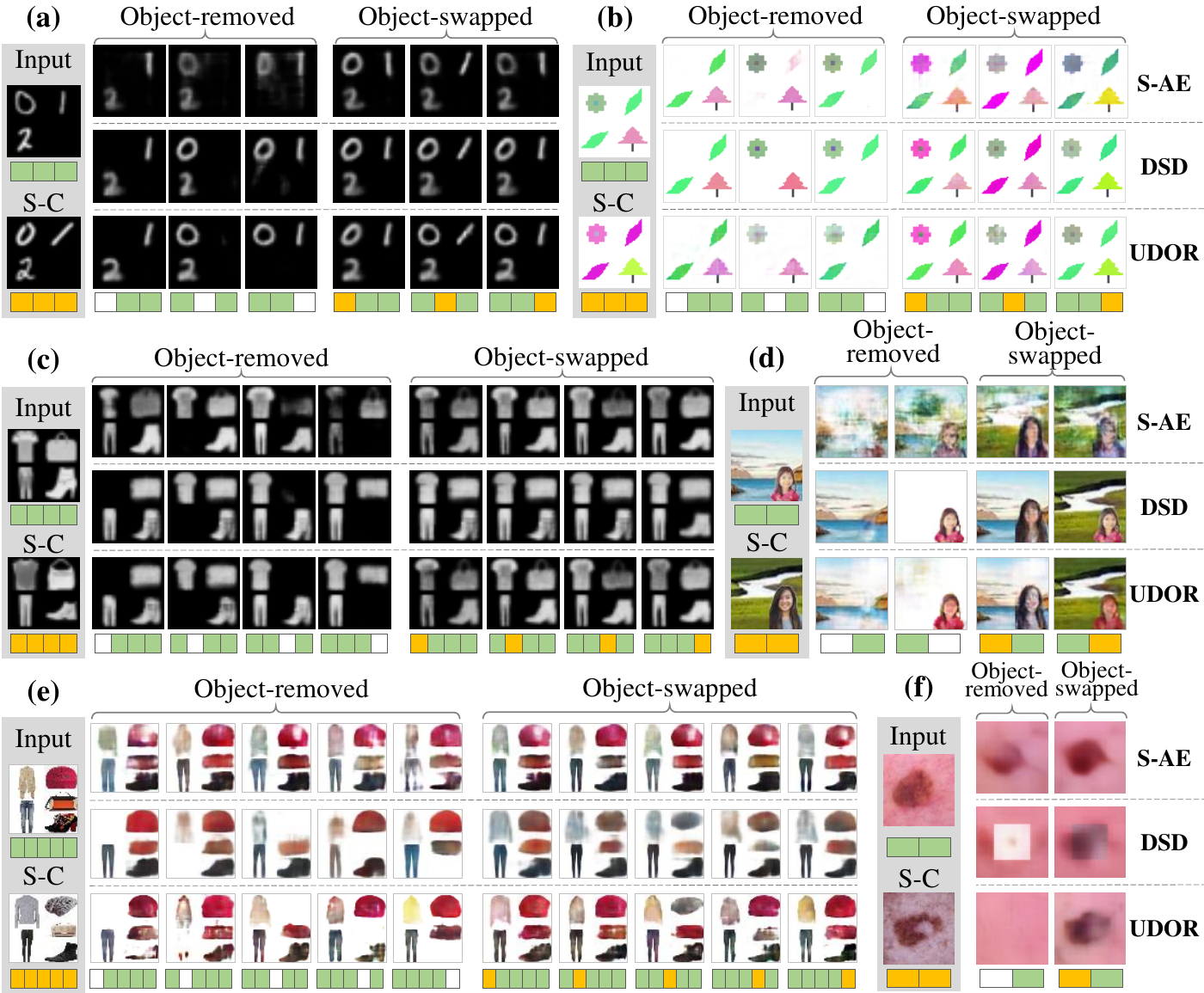}
\caption{The qualitative results of S-AE (Section \ref{Qualitative Evaluation}), DSD \cite{Feng2018Dual} and UDOR on six datasets: (a) Multi-MNIST, (b) Pattern-Design, (c)Multi-Fashion, (d) Mugshot, (e) Outfit, (f) HAM. For all images, the corresponding latent representations are given under images. `Object-removed' denotes resetting each part of representation as empty vector ( denoted by the `white rectangle') by turns. `Object-swapped' denotes swapping each part of representation of the swapping candidate image (S-C) into the representation of input image.}
\label{visualResults}
\end{figure}

\subsection{Qualitative Evaluation}
\label{Qualitative Evaluation}
As described above, there are few disassembling object representation methods so far. So we compare our unsupervised method with two most related (semi-)supervised methods, which can validate the performance of the UDOR intuitively. The first compared supervised method is a Supervised Auto-Encoder (S-AE), which adopts the annotated object labels to supervise each part of the representation learned by basic auto-encoder through a classifier. The encoder, decoder, and classifier have the same network architecture as UDOR's. The detail network architecture of S-AE is given in the supplementary material A. What's more, the UDOR is also compared with the semi-supervised method DSD~\cite{Feng2018Dual}, which can be transferred into the object representation disassembling method by replacing the annotated attribute input with annotated object samples.

Fig.~\ref{visualResults} gives some visualization results of the above methods on the six datasets. For each dataset, we show an input sample, a swapping candidate image (S-C). Meanwhile, we demonstrate the object-removed images and object-swapped images, which compares the disassembling performance of different methods qualitatively. The object-removed images are reconstructed through resetting different parts of representation to an empty vector, which is a part of the empty image's representation. The object-swapped images are reconstructed by swapping part of the swapping candidate image's representation into the representation of the input image.

From Fig.~\ref{visualResults} (d-f), we see that S-AE fails to remove corresponding objects from the object-removed results, which demonstrates that only class labels are not enough for directly disassembling representation on complicated color datasets.
What's more, S-AE achieves the worst visual results. The reason is that the label only supervises the S-AE extract relevant features of the corresponding object into each part of the representation. However, it cannot prevent each part from extracting irrelevant features. The Object-removing Operation of our method and swapping module of DSD can restrict irrelevant features of other objects to be extracted into the corresponding part.
From Fig.~\ref{visualResults} (a-f),  we can see that corresponding objects are successfully removed from the results of DSD and UDOR. On the average, DSD achieves the best visual results on the above six datasets. However, the results of UDOR are sharper and clearer than other methods. It should be noted that the paired samples of DSD are generated with hand-cut center patches, which leads to the mosaic results in Fig.~\ref{visualResults} (f).
In sum total, the proposed UDOR, despited unsupervised, achieves truly encouraging results on par with those of (semi-)supervised methods.
More visual results are provided in the supplementary material C.

\begin{figure}[!t]
\centering
\includegraphics[scale =0.415]{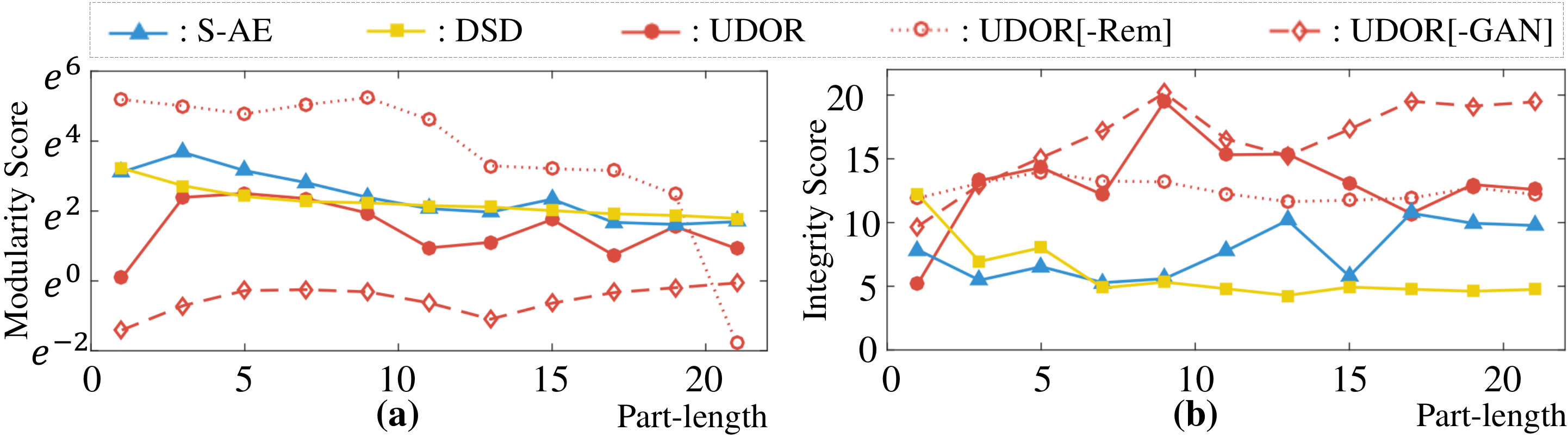}
\caption{The \emph{modularity scores} and \emph{integrity scores} of different methods in different part-lengths. \emph{Part-length} is the length of the representation's part. The S-AE is the Supervised Auto-Encoder (Section ~\ref{Qualitative Evaluation}). The DSD is the  semi-supervised method~\cite{Feng2018Dual}. UDOR[-Rem] and UDOR[-GAN] denote UDOR without object-removing loss $\mathbf{\mathcal{L}_{rem}}$ and WGAN-GP loss $\mathbf{\mathcal{L}_{adv}}$, respectively. `$e$' is the mathematical constant.}
\label{ModularityandIntegrity}
\end{figure}

\subsection{Quantitative Evaluation}
To evaluate the disassembling performance quantitatively, we compare the proposed UDOR with S-AE and DSD~\cite{Feng2018Dual} on the Multi-MNIST using the \emph{modularity score} and \emph{integrity score} in different part-lengths.
In the experiment, $T$ and $D$ are set to $300$ and $10$, respectively. We sample $11$ part-lengths ($\{1, 3, 5, ..., 21\}$) and test all methods in those part-lengths setting.

Fig.~\ref{ModularityandIntegrity} plots the modularity scores and integrity scores of different methods in different part lengths. In Fig.~\ref{ModularityandIntegrity}(a), the modularity scores of UDOR are smaller than DSD's and S-AE's in almost all part-lengths, which demonstrates that our unsupervised method achieves better disassembling performance than (semi-)supervised methods (S-AE and DSD) on modularity. The primary cause is that the Object-removing Operation can reduce the correlation between the parts of representation effectively. It can be verified by the modularity score of UDOR[-Rem], which is the highest than other methods in almost all part-lengths. It means that UDOR without object-removing loss achieves the worst modularity performance on the disassembled object representation.

For the integrity score, UDOR achieves a smaller score than S-AE and DSD in part-length $1$, which is shown in Fig.~\ref{ModularityandIntegrity}(b). However, S-AE and DSD achieve smaller scores than our method in large part-lengths, which means that the (semi-)supervised method achieves better visual performance in large part-length. In general, each part of representation tends to contain more irrelevant features of other objects along with the part-length increases. For those (semi-)supervised methods, the label will supervise each part containing more relevant features of the corresponding object, which leads to better visual results. In summary, our proposed unsupervised method achieves better modularity performance in almost all part-lengths and achieves better visual results in part-length $1$ than (semi-)supervised methods (S-AE and DSD).

\subsection{Ablation Study}
In the UDOR, the total loss $\mathbf{\mathcal{L}}$ (Eqn. (\ref{eq5})) is composed of four parts: the basic reconstruction loss $\mathbf{\mathcal{L}_{rec}}$, the fuzzy classification loss $\mathbf{\mathcal{L}_{cla}}$, the object-removing loss $\mathbf{\mathcal{L}_{rem}}$, and the generative adversarial loss $\mathbf{\mathcal{L}_{adv}}$. The $\mathbf{\mathcal{L}_{rec}}$ (Eqn. (\ref{eq1})) is the basic reconstruction loss of double AE, which ensures that all the features of the image are encoded into the latent representation entirely. The fuzzy classification loss $\mathbf{\mathcal{L}_{cla}}$ (Eqn. (\ref{eq2})) is the core component for disassembling. Without the $\mathbf{\mathcal{L}_{rec}}$ and $\mathbf{\mathcal{L}_{cla}}$, our framework will fail in disassembling object representation. The object-removing loss $\mathbf{\mathcal{L}_{rem}}$ (Eqn. (\ref{eq4})) and generative adversarial loss $\mathbf{\mathcal{L}_{adv}}$ (Eqn. (\ref{eq3})) is devised to improve the modularity of latent representation and visual integrity of reconstructed image, respectively.

We then conduct the ablation study by removing the $\mathbf{\mathcal{L}_{rem}}$ and the $\mathbf{\mathcal{L}_{adv}}$ from the framework.
Fig.~\ref{ModularityandIntegrity} gives the modularity and integrity scores of the methods without the object-removing loss $\mathbf{\mathcal{L}_{rem}}$ (UDOR[-Rem]) and the generative adversarial loss $\mathbf{\mathcal{L}_{adv}}$ (UDOR[-GAN]). In the Fig.~\ref{ModularityandIntegrity}(a), UDOR[-Rem] achieves the larger modularity score than UDOR, which demonstrates the effectiveness of object-removing loss for improving the modularity of latent representation. However, UDOR[-GAN] has a smaller modularity score than UDOR, which shows that the $\mathbf{\mathcal{L}_{adv}}$ affects the modularity of latent representation negatively. From Fig.~\ref{ModularityandIntegrity}(b), we can see that UDOR[-GAN] has the larger scores than UDOR and UDOR[-Rem], which verifies that WGAN-GP module can effectively improve the visual integrity of reconstructed images.

\begin{figure}[!t]
\centering
\includegraphics[scale =0.77]{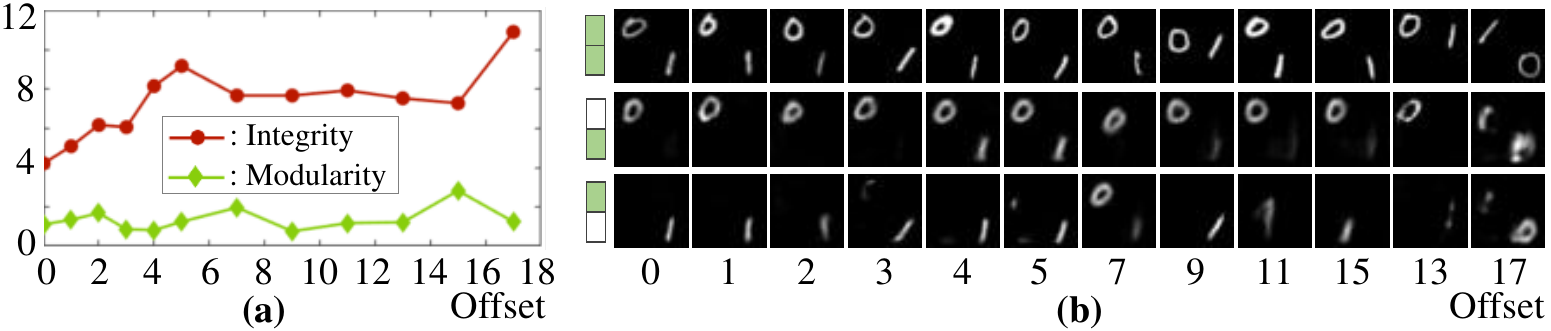}
\caption{The results of our method on different object position offsets. a) The integrity score and modularity score along with different offsets. b) The first row is the original input images, and the next two rows are reconstructed images with the representations, the first and second part of which are reset as empty parts( the `white rectangle').}
\label{different_offset}
\end{figure}

\subsection{Object Position}
\label{Object_Position}
The object position is an important factor for disassembling object representation. We generate 12 different datasets with different offset settings of object position ($\{0, 1, 2, 3, 4, 5, 7, 9, 11, 13, 15, 17\}$). In the dataset, each sample is composed of one/two digit objects. All the samples are $32\times32$ images. With those datasets, we train 12 models with $10000$ samples and test them with $2000$ test samples. The corresponding quantitative and qualitative results are shown in Fig.~\ref{different_offset}. The integrity scores and modularity scores along different offsets are shown in Fig.~\ref{different_offset}(a), where we can see that the integrity score becomes large when the offset is large than $3$. The corresponding object-removed images are shown in Fig.~\ref{different_offset}(b) on different offset setting datasets. Like the integrity score, the corresponding objects can't be removed entirely when the offset is large than $3$. In summary, UDOR is robust on small object position offset and fails on large offset variance, which is the primary direction of our future research.

\begin{table}[!t]
\scriptsize
\begin{center}
\begin{tabular}{|l|cccccc|cccccc|}
\hline
\multirow{2}{*}{\bf Method}  & \multicolumn{6}{c|}{\bf Muti-MNIST dataset} & \multicolumn{6}{|c|}{\bf HAM dataset ~\cite{tschandl2018the}}\\
\cline{2-13}
{}        &  C-P  &  C-R  &mic-F1 &O-P    &O-R  &macro-F1 & C-P   &C-R   &mic-F1  & O-P   & O-R   &macro-F1\\
\hline
{\bf S-AE}      & 100   & 100   & 100   & 100   & 100   & 100   & 38.69 & 27.26 & 31.58 & 69.80 & 69.80 & 69.80\\
{\bf DSD}       & 95.86 & 98.18 & 96.81 & 95.75 & 98.18 & 96.83 & 39.72 & 29.87 & 28.49 & 68.60 & 68.60 & 68.60\\
{\bf UDOR}      & 99.95 & 99.45 & 99.70 & 99.95 & 99.46 & 99.70 & 55.16 & 34.03 & 44.03 & 75.53 & 75.53 & 75.53\\
{\bf UDOR[-Rem]} & 98.36 & 92.92 & 95.09 & 98.16 & 92.97 & 94.32 & 47.41 & 26.58 & 36.90 & 70.91 & 70.91 & 70.91\\
{\bf UDOR[-GAN]} & 87.49 & 93.49 & 88.95 & 87.56 & 93.49 & 88.87 & 38.76 & 24.92 & 32.72 & 65.32 & 65.32 & 65.32\\

\hline
\end{tabular}
\end{center}
\caption{The classification performance of different methods (All scores in $\%$). `C-P' and `O-P' denote per-class and overall precision scores. `C-R' and `O-R' denotes per-class and overall recall scores. `micro-F1' and `macro-F1' scores \cite{tang2009large} describe the geometrical average of the precision and recall scores.  }
\label{table:score1}
\end{table}

\subsection{Applications}
As described above, our method can be applied to many machine learning tasks, including image editing,  few- or zero-shot learning, classification, and so on. In this section, we test the performance on two basic applications: image editing and classification. For image editing, Fig.~\ref{visualResults} gives the object-removed and object-swapped results on six datasets. In the object-removed results, the corresponding objects are effectively removed, while the corresponding objects are swapped in the object-swapped results. It validates that our method can be well applied in the editing of many image scenes. More editing results can be found in the supplementary material C.

For the classification, we compare the UDOR with S-AE and DSD on Multi-MNIST and HAM~\cite{tschandl2018the}.
After getting the disassembled representations, we adopt a simple linear SVM (Details are given in the supplementary material A) to train and test the classification performance. The simple classifier can reduce its influence on the final classification performance, which can help better measure the classification effect of different object representations. It should be noted that the classification performance can be promoted with more powerful classifiers and feature extractors. From Table~\ref{table:score1}, we can see that UDOR achieves a higher score than DSD's on Muti-MNIST dataset and achieves the best performance on HAM dataset~\cite{tschandl2018the}, which demonstrates that the object features extract by our method is more intact and independent. The S-AE achieves the best and the worst performance on the Multi-MNIST dataset and HAM dataset, which demonstrates that class labels are only directly effective for simple datasets.
Meanwhile, the UDOR[-Rem] and UDOR[-GAN] achieve the lower scores than UDOR, which verifies the effectiveness of the Object-removing Operation and the WGAN-GP module once again. It's noticeable that the UDOR[-GAN] achieves the worst performance on the classification task, which indicates that the WGAN-GP module can effectively enhance the integrity and independence of the object's features.

\subsection{Failure Case and Future Works}
The above experiments show that the UDOR can achieve closed disassembling performance as (semi-)supervised methods. However, there are still some shortcomings. Firstly, there are many balance parameters in total loss (Eqn. (\ref{eq5})). For different datasets, the balance parameters need to be fine-tuned. Ill-suited settings may lead to overfitting or non-convergence. The detail balance parameters for all the datasets can be found in supplementary material B.

\begin{figure}[!t]
\centering
\includegraphics[scale =0.815]{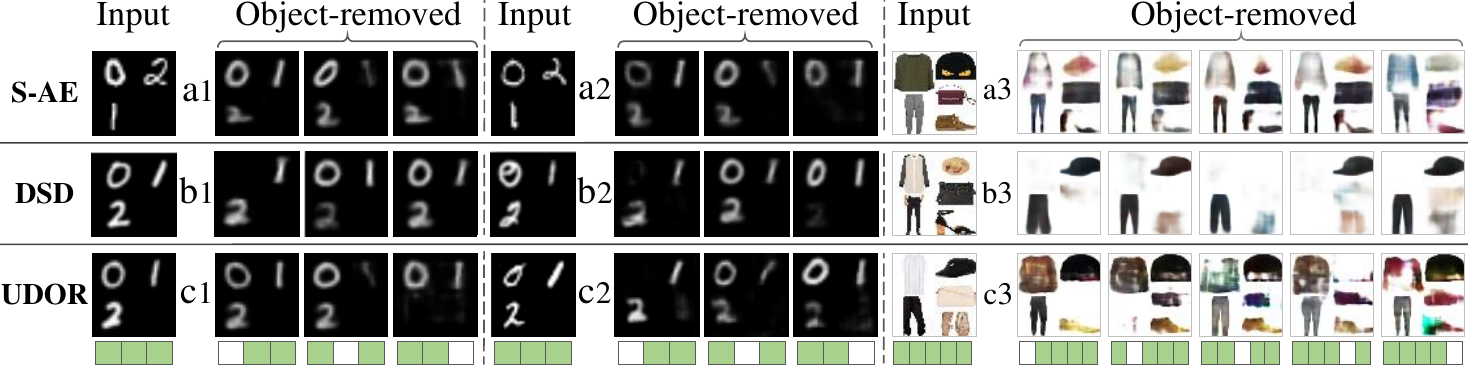}
\caption{Some failure cases on the Multi-MINIST dataset and the Outfit dataset.}
\label{fail_case}
\end{figure}

Some failure cases are shown in Fig.~\ref{fail_case}, where the digit zero is failed to be entirely disassembled in the case `c1' of UDOR. The failed cause is that the model converges to a local minimum. In case `c2', digit one and digit two are not well disassembled, which leads to the ghost of those digits in the reconstructed images. It is noticeable that those failure case also occurs in the S-AE and DSD.
The primary cause of the ghost is the stability of converging in the training stage, which is a major direction in our future research. From Fig.~\ref{fail_case}(b3,c3), we can see that failure cases of DSD and S-AE are more serious when handling the complicated outfit dataset. Meanwhile, the S-AE fails to disassemble object representations on complicated image datasets, which is illustrated in Fig.~\ref{fail_case}(a3) and Fig.~\ref{visualResults}(d,e,f).
In Section~\ref{different_offset}, the experiments on object position offset are expounded in detail, which demonstrates that the UDOR will fail on large offset variance. Therefore, combining position information into the disassembled representation is a significant direction in our future research.

\section{Conclusion}
In this paper, we study a new representation-learning task, termed {disassembling object representation}, for which the goal is to disassemble features of different objects within an image into
distinct parts of a latent representation.
Towards solving this task,  we propose an unsupervised strategy,
termed Unsupervised Disassembling Object Representation~(UDOR).
UDOR follows a double AE architecture,
in which a Fuzzy Classification and a
Object-removing Operation are imposed to
achieve modularity and visual integrity.
Furthermore, we devise two disassembling metrics
to measure the modularity of representations and the integrity of images, respectively. Experiments on
several datasets
show that the proposed UDOR accomplishes favorable results,
on par with those of supervised methods.
In future work, we will focus on the stabilized convergence of UDOR and large changes of object's position.
\bibliographystyle{splncs04}
\bibliography{egbib}
\end{document}